\pdfoutput=1

\documentclass[11pt]{article}

\usepackage[]{acl}
\usepackage{tcolorbox}

\usepackage{times}
\usepackage{latexsym}

\usepackage[T1]{fontenc}

\usepackage[utf8]{inputenc}

\usepackage{microtype}

\usepackage{inconsolata}

\usepackage{inconsolata}
\usepackage{mathtools}  
\usepackage{adjustbox}
\usepackage{booktabs}
\usepackage{fancyvrb}
\usepackage{xcolor}
\usepackage{amsfonts}
\usepackage{xspace}
\newcommand{\steerlm}{\textsc{SteerLM}\xspace}
\newcommand{\helpsteer}{\textsc{HelpSteer}\xspace}
\usepackage{cleveref}

\title{\helpsteer: Multi-attribute Helpfulness Dataset for \steerlm}

\author{Zhilin Wang, Yi Dong, Jiaqi Zeng, Virginia Adams, \\
\textbf{Makesh Narsimhan Sreedhar, Daniel Egert, Olivier Delalleau,} \\
\textbf{Jane Polak Scowcroft, Neel Kant, Aidan Swope, Oleksii Kuchaiev} \\
  NVIDIA \\
  \texttt{\{zhilinw, yidong\}@nvidia.com} 
  }

\begin{document}
\maketitle
\begin{abstract}

Existing open-source helpfulness preference datasets do not specify what makes some responses more helpful and others less so. Models trained on these datasets can incidentally learn to model dataset artifacts (\textit{e.g.}  preferring longer but unhelpful responses only due to their length). To alleviate this problem, we collect \helpsteer, a multi-attribute helpfulness dataset annotated for the various aspects that make responses helpful. Specifically, our 37k-sample dataset has annotations for correctness, coherence, complexity, and verbosity in addition to overall helpfulness of responses. Training Llama 2 70B using the \helpsteer dataset with \steerlm technique produces a model that scores 7.54 on MT Bench, which is currently the highest score for open models that do not require training data from more powerful models (\textit{e.g.} GPT4). 
We release this dataset with CC-BY-4.0 license at \url{https://huggingface.co/datasets/nvidia/HelpSteer}

\end{abstract}

\section{Introduction}

\begin{figure}[t]
    \centering
    \includegraphics[width=7.6cm]{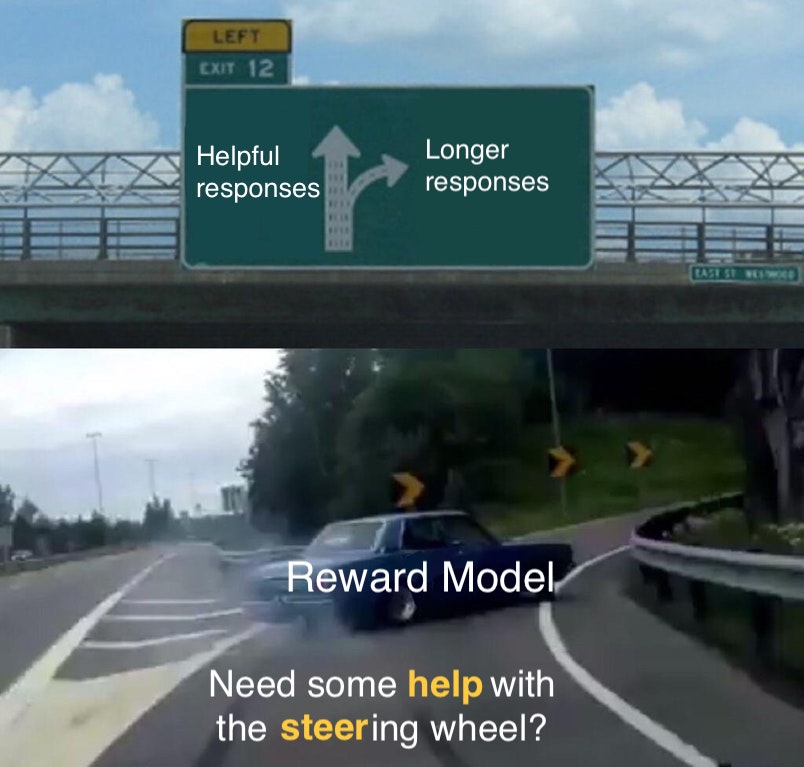}
    \caption{\helpsteer can provide multi-attribute helpfulness information for modeling human feedback, reducing the likelihood of models learning dataset artifacts such as preferring responses simply due to their length.}
    \label{fig:front_page}
\end{figure}

Helpfulness and Safety are the twin prime objectives for aligning domain-general models to follow user instructions \citep{ouyang2022training, bai2022training, touvron2023llama}. While various studies \citep{ganguli2022red, ji2023beavertails, rebedea2023nemo} have thoroughly defined the aspects of safety, the criteria for what constitutes a \textit{helpful} model remains opaque. Recent trends have suggested a shift from asking this question altogether and settling on the notion that model responses should align with user or annotator preferences. While using black-box user/annotator preferences can guide us towards more helpful model responses, this approach is both intellectually unsatisfying and inefficient in terms of compute and the amount of data required \citep{touvron2023llama, bai2022training}. The use of a black-box approach also leads to the possibility for models to associate longer responses with helpfulness, thereby prioritizing length over substance \citep{dong2023steerlm, dubois2023alpacafarm, singhal2023longway}. This poses the risk of valuing verbose but irrelevant responses, highlighting the need for a more refined understanding of `helpfulness' in model training.

Training more effective domain-general language models requires elucidating what humans find helpful in responses. \citet{kopf2023openassistant} use the aspects of creativity and humor as a measure of the helpfulness of instruction-following models. While these attributes can enhance the utility of responses in specific settings (\textit{e.g.} story writing), they do not contribute to helpfulness in other scenarios (\textit{e.g.} text classification) and could even be counterproductive in formal business settings. Insights from domain-specific language models trained for particular tasks offer valuable guidance on what constitutes helpfulness. In summarization tasks, \citet{10.5555/3495724.3495977} identify accuracy, coverage, and coherence as important components of the overall quality of summaries. Similarly, \citet{wu2023finegrained} emphasize the importance of relevance, factuality, and information completeness in the long-form question-answering task. These findings suggest that the factors contributing to helpfulness in language models may vary significantly across different applications and contexts.

We propose that the helpfulness of domain-general model responses can be assessed by their correctness, coherence, complexity, and verbosity. \textbf{Correctness} refers to the inclusion of all pertinent facts without errors. \textbf{Coherence} refers to the consistency and clarity of expression. We believe both aspects are critical for users to trust model responses across all tasks. \textbf{Complexity} represents the response's intellectual depth, reflecting whether the content is basic or requires profound expertise (\textit{i.e.} whether the response can be written by anyone with basic language competency or requires deep domain expertise to author), and is important because people tend to trust experts more. \textbf{Verbosity} refers to the amount of detail included in the response, which has been found to correlate positively with human preferences \citep{dong2023steerlm, dubois2023alpacafarm}, potentially because lengthy responses likely contain additional relevant information. While not exhaustive, we believe this set of factors provides a foundational framework for curating data to enhance the helpfulness of language models.

To demonstrate the contributions of these attributes, we: 

\begin{enumerate}
    \item Curate a helpfulness dataset with 37k conversations, with each response annotated for correctness, coherence, complexity, and verbosity in addition to overall helpfulness.
    \item Leverage this helpfulness dataset to align  a model which has the highest MT Bench score of 7.54 among models that do not require training data from powerful proprietary models (\textit{e.g.} GPT4).
    \item Openly release this resulting dataset with a CC-BY-4.0 license to enable the community to build upon our findings.
\end{enumerate}

\section{Related Works}

\begin{table*}[ht!]
\centering
\begin{adjustbox}{max width=\textwidth, scale=1
}
\begin{tabular}{lccccc}
\toprule
\textit{Name} & \textit{Helpfulness-relevant Attributes} & \textit{N conv. (k)} & \multicolumn{2}{c}{Mean Length in chars (Std.)} \\
&&& Prompt & Response \\

\midrule
\helpsteer &  Helpfulness, Correctness, Coherence, Complexity, Verbosity & 37.1 & 2491.8 (1701.7) & 497.3 (426.7) \\
Open Assistant  & Quality, Creativity, Humor &  59.4 & 397.5 (620.8) &396.2 (618.8)\\
HH RLHF &  - &   337.7 &  794.4 (706.9) & 310.7 (311.4)\\

\bottomrule
\end{tabular}
\end{adjustbox}
\caption{Overview of Open-source Helpfulness Preference Modeling Datasets}
\label{tab:overview_preference}
\end{table*}

\paragraph{Helpfulness Preference Datasets}

An overview of open-sourced domain-general helpfulness preference datasets can be found in Table \ref{tab:overview_preference}. HH-RLHF \citep{bai2022training} is a popular ranking-based dataset containing pairs of responses, one of which is the preferred response and the other is the rejected one. In addition to ranking data, Open Assistant \citep{kopf2023openassistant} also contains helpfulness-relevant attributes labeled for each response. Specifically, annotators were asked to rate each response for their quality, creativity and humor on a 5-point likert scale, which can then be useful for training \steerlm models \citep{dong2023steerlm}.

Contemporary works \citep{sharma2023understanding, cui2023ultrafeedback} have also made use of GPT-4 to annotate for various aspects contributing to helpfulness such as truthfulness and the ability to follow instructions. Given the lack of clarity on the biases and limitations of GPT-4 in performing such annotations \citep{cui2023ultrafeedback}, we find it difficult to trust such automated annotations, especially when subsequent human validation of these annotations are not done. An additional risk lies in the OpenAI GPT-4 Terms of Use \footnote{\url{https://openai.com/policies/terms-of-use}}, specifically
\begin{tcolorbox}
Section 2 (c) Restrictions. You may not ... (iii) use output from the Services to develop models that compete with OpenAI.
\end{tcolorbox}
This makes training models using GPT-4 output potentially litigious and puts into question whether GPT-4 annotated data can \textit{truly} be open-sourced.

Additionally, there are several helpfulness preference datasets for specific tasks/domains. These include summarization \citep{10.5555/3495724.3495977}, question answering \citep {nakano2022webgpt, wu2023finegrained}, solving code problems \citep{h4stackexchange} as well as Reddit conversations \citep{pmlr-v162-ethayarajh22a, wang-torres-2022-helpful}. Given the task/domain-specificity of these datasets, they are unlikely to improve performance across a diversity of tasks, which is the purpose of \helpsteer.

\section{Dataset}
In this section, we present details on the data collection methodology of \helpsteer outlining the underlying motivation, the prompt selection strategy, the procedure for generating responses, and the measures implemented to ensure the quality of annotations.

\subsection{Data collection}

\paragraph{Motivation}

We collect \helpsteer based on the limitations we encountered when using the Open Assistant dataset with \steerlm technique \citep{dong2023steerlm}. Although the responses were largely helpful, some instances revealed issues such as factual inaccuracy, incoherence, oversimplification, or excessive verbosity. Additionally, we observed suboptimal performance on certain tasks like Rewrite, Summarization, Classification, Extraction, and Closed Question Answering. These tasks typically involve a reference text, which may be less appealing to volunteers in the Open Assistant project due to the increased length and complexity of the prompts. To obtain better annotations for attributes contributing to helpfulness of responses for such tasks, we used a substantial number of prompts from these categories and collected annotations for correctness, coherence, complexity, and verbosity in addition to overall helpfulness to enhance the dataset's utility.

\paragraph{Prompt Collection}

We first collect 10,459 single-turn prompts. Approximately half of these were created by Scale AI, an external data annotation vendor, while the remainder were synthetically generated using templates to ensure diversity in prompt sources.  Initially, a larger set of prompts was produced, from which we filtered out about 20\% deemed unsatisfactory. Following \citet{ouyang2022training}, our collection included prompts from Open Question Answering, Generation, and Brainstorming tasks, along with the five tasks (Rewrite, Summarization, Classification, Extraction, and Closed Question Answering) that had limited representation in the Open Assistant dataset. We maintained a roughly 10\% distribution for each category, with the exception of the five tasks where \steerlm showed diminished performance, to which we allocated the remaining 20\%.

\paragraph{Response Generation} 

To generate responses, we utilized our in-house 43 billion parameter model, producing four distinct responses for each prompt. These responses were generated within the maximum context length of 4,096 tokens. Our configuration involved a temperature setting of 1.0 and a top\_p value of 0.80, coupled with a repetition penalty of 1.0 and a top\_k setting of 1000 resulting in diverse yet reasonable responses.

\paragraph{Response Annotation}

Each response in our dataset was evaluated based on five attributes: Helpfulness, Correctness, Coherence, Complexity, and Verbosity. These were rated on a Likert-5 scale, ranging from 0 to 4. Unlike annotations for RLHF \citep{bai2022training, ouyang2022training, touvron2023llama} which involve comparison with other responses to the same prompt, each response was rated independently of other responses. We found this approach to be more scalable than RLHF annotations as its comparative model results in quadratic growth in annotation workload relative to the number of responses per prompt, while our approach (\helpsteer) increases linearly.

For \helpsteer annotations, we engaged a select group of contractors via Scale AI. These contractors were provided with comprehensive guidelines that defined each attribute and the criteria for every rating level, together with some annotated examples. These guidelines and examples are detailed in Appendix \ref{sec:appendix_annotation_guidelines}.

The annotation process involved approximately 200 U.S.-based human annotators. Candidates first underwent preliminary assignments, including assessments of English proficiency, to determine eligibility for working on the \helpsteer project. Subsequently, they participated in an introductory training course on the task which ended with a test that involved annotating 35 sample responses. This process ensured not only a thorough understanding of the task requirements but also the delivery of high-quality annotations.

Post-annotations, Scale AI performed extensive quality assurance, with each annotation reaching a minimum of two human reviews in addition to automated checks. After receiving the annotations from Scale AI, we conducted our independent quality assurance to make sure that the quality of the annotations was up to our expectations. After filtering out annotations that did not meet our criteria at various stages, we finalized a dataset consisting of 37,120 high-quality annotated samples. 

\subsection{Data Analysis}

\begin{table}[ht!]
\centering
\begin{adjustbox}{max width=\columnwidth, scale=1
}
\begin{tabular}{lccc}
\toprule
\textit{Attribute} & \textit{Pearson R} &  \textit{Mean} & \textit{Std.}\\
&w. helpfulness & \\
&/ quality & \\

\midrule
\multicolumn{2}{l}{\textbf{\helpsteer}} \\
\midrule
helpfulness & 1 & 2.7856 & 0.9793 \\
correctness & 0.8525 & 2.8369 & 0.9935\\
coherence & 0.6348 & 3.2991 & 0.7699\\
complexity & 0.2361 & 1.4423 & 0.8205\\
verbosity & 0.2555 & 1.5331 & 0.9287\\ 

\midrule
\multicolumn{2}{l}{\textbf{Open Assistant}}\\
\midrule
quality & 1 & 2.5735 & 0.9878 \\
creativity & 0.3428 & 1.5764 & 1.0618 \\
humor & -0.0992 & 0.7218 & 0.8507 \\

\bottomrule
\end{tabular}
\end{adjustbox}
\caption{Descriptive statistics for helpfulness-relevant attributes in \helpsteer and Open Assistant. In Open Assistant, the attribute of Quality most closely resembles our definition of helpfulness. Scores for each attribute are between 0 and 4 on a Likert-5 scale.}
\label{tab:descriptive_attributes}
\end{table}

With 37.1k conversations, \helpsteer is comparable in size to the 59.4k conversations found in the Open Assistant \citep{kopf2023openassistant} dataset which are annotated with attributes contributing to helpfulness. Prompts found in \helpsteer have a mean length of 2491.8 characters (std. = 1701.7) while responses have mean length of 497.3 characters (std. = 426.7). This means that \helpsteer prompts are substantially longer than those in Open Assistant (397.5 characters with std. = 620.8) and this can be primarily attributed to the inclusion of tasks like Summarization, Rewrite, and Extraction, which incorporate reference texts within the prompts.

Table \ref{tab:descriptive_attributes} provides a detailed breakdown of the distribution of each attribute and their Pearson correlation with helpfulness in \helpsteer, and quality in Open Assistant. In \helpsteer, correctness and coherence exhibit a strong positive correlation with helpfulness (Pearson's $R > 0.6$) while complexity and verbosity are weakly correlated (Pearson's $R > 0.2$).  This suggests that the correctness and coherence of a model response have a significant influence on perceived helpfulness, while verbosity and complexity are less pivotal. The attributes in Open Assistant are either weakly correlated (Creativity with Pearson's $R = 0.34$) or slightly negatively correlated (Humor with Pearson's $R = -0.099$). This aligns with our hypothesis that creativity and humor, while potentially enhancing, are not essential for helpfulness and can sometimes detract from it.

\helpsteer's attribute distribution indicates that responses typically exhibit high coherence (average 3.30 out of 4), moderate correctness (average 2.84), and are relatively low in complexity (average 1.44) and verbosity (average 1.53) together resulting in moderately helpful (2.78) responses. Responses in Open Assistant are similar in overall quality (2.57), but they are low in both creativity (1.58) and humor (0.72), suggesting that these attributes are insufficient to explain the overall helpfulness as supported by their low/negative Pearson correlation scores (Pearson's $R < 0.35$) with helpfulness.

An Ordinary Least Squares Regression analysis, with helpfulness as the dependent variable and the four attributes as independent variables in \helpsteer, revealed significant contributions of each attribute to overall helpfulness ($p<0.05$). Collectively, these attributes account for a considerable 73.0\% of the variance in helpfulness, offering a comprehensive understanding of the factors driving helpfulness in model responses.

\section{Experiments}

In this section, we detail our methodology in assessing whether \helpsteer can effectively guide models towards improved helpfulness, factual accuracy, coherence, and appropriate levels of complexity and verbosity. We employ a blend of automated and human evaluations to gauge these aspects. Additionally, we describe the process of training a \steerlm model \citep{dong2023steerlm} utilizing the \helpsteer dataset. The section also covers the baseline models against which we compare the performance of the \steerlm model, providing a comprehensive view of its effectiveness in model alignment.

\subsection{Automatic Evaluation}

\paragraph{Helpfulness} We follow \citet{jiang2023mistral, lu2023instag} to use MT Bench \citep{zheng2023judging} for helpfulness evaluation. MT Bench consists of 80 multi-turn questions, each consisting of an initial question and a follow-up question, for a total of 160 prompts. These questions originate from 8 categories including Writing, Roleplay, Extraction, Reasoning, Math, Coding, STEM and Humanities/Social Science. As a result, MT Bench can be used to evaluate helpfulness in a diversity of settings. We first 
greedily generate responses with up to 1024 tokens (default value for MT Bench). The responses to these prompts are evaluated by GPT-4 to give a score between 1 and 10, and we report the mean across all prompts with a higher MT Bench score indicative of greater helpfulness.

\paragraph{Correctness} We follow \citet{ouyang2022training, bai2022training, touvron2023llama} in using TruthfulQA \citep{lin-etal-2022-truthfulqa} to evaluate factuality of models. TruthfulQA consists of 817 questions across 38 categories (\textit{e.g.} health, finance and legal). We use TruthfulQA MC2 as used in the Huggingface OpenLLM Leaderboard\footnote{\url{https://huggingface.co/spaces/HuggingFaceH4/open_llm_leaderboard}}, which represents the normalized total probability assigned to the set of one or more true answers out of 4 to 5 answer options per question. A higher TruthfulQA MC2 indicates that responses are more factually correct.

\paragraph{Coherence} We follow \citet{laban-etal-2021-transformer} in using Base Language Model Perplexity as a measure of text coherence.  For this measure, we calculate the perplexity of MT Bench responses using evaluation code from the original authors\footnote{\url{https://github.com/tushar117/Transformer-Models-for-Text-Coherence-Assessment}}.
While the best performing non-finetuned model in \citet{laban-etal-2021-transformer} was the GPT2-medium model, we found it to be unsuitable for our use case because perplexity was more than 100 for many responses suggesting that GPT2-medium was undertrained. Instead, we use the Llama 2 13B Foundation model perplexity, which we show to have higher accuracy on the Zero-Shot Shuffle Test\footnote{For Llama 2 13B vs. GPT2-medium - Legal: 99.7 vs 98.6; Reddit 98.5 vs 88.9; WSJ corpus was inaccessible behind a paywall with ref. to Table 1 of \citet{laban-etal-2021-transformer}}, as used in \citet{laban-etal-2021-transformer}. A lower perplexity implies that responses are more coherent.

\paragraph{Complexity} We follow \citet{scarton-specia-2018-learning} and \citet{scialom2021rethinking} to use Flesch-Kincaid Grade Level (FGKL) as a metric for text complexity \citep{Kincaid1975DerivationON}. FKGL represents the US grade level (\textit{i.e.} 1 to 12 where 12 is the last year of pre-university education) which the text is targetted at. We calculate FKGL based on MT Bench responses using the Easse package \citep{alva-manchego-etal-2019-easse}. Higher FKGL means higher text complexity.

\paragraph{Verbosity} We use the mean number of characters in MT Bench responses as a measure for verbosity.

\subsection{Human Evaluation}
 Following \citet{dong2023steerlm}, we conduct human evaluations to assess the relative helpfulness of model responses to complement automatic evaluation.

\paragraph{Data}
We select the first-turn prompts from the MT Bench dataset, comprising 80 open-ended questions on diverse topics including math, science, coding, roleplaying, reasoning etc.

\paragraph{Annotation Process}

We recruited 12 volunteers with at least undergraduate education in computer science or related fields to evaluate the quality of model responses in a blind setting. Annotators were presented with the prompt and 3 model responses in random order, and asked to rank the responses in the order of helpfulness. To reduce annotator fatigue, prompts were divided into 4 sets of 20 prompts, and each annotator was assigned one set. This means that prompts to every response was independently ranked by 3 annotators (Fleiss' $\kappa$=0.383).

\paragraph{Metrics}
We show the win rate of models against one another based on pairwise model response preferences from human evaluations. Additionally, we calculate an Elo score following \citet{vicuna2023} to better illustrate how models compare to one another. We begin with an initial score of 1000 and $K=32$ and repeat the procedure 10000 times to account for the ordering effect in calculating Elo scores.

\subsection{Foundation models} We use the Llama 2 Foundation models \citep{touvron2023llama} for all our experiments - the 70B variant for the main language model and the 13B variant as the Attribute Prediction Model and Reward Model in \steerlm and RLHF baseline respectively. Our initial explorations showed little benefit in using a larger model (\textit{i.e.} the 70B model) and large increases in compute requirements.

\subsection{\steerlm}
We train the Llama 2 13/70B model following the \steerlm approach \citep{dong2023steerlm}. \steerlm is a model alignment method (alternative to RLHF) with four key steps. First, an Attribute Prediction Model is trained to predict scores for multiple semantic attributes that capture dimensions of response helpfulness such as correctness and coherence. Next, datasets consisting of prompt-response pairs are annotated with these attributes using the Attribute Prediction Model. Then, Attribute Conditioned Supervised Fine-Tuning (AC-SFT) is performed by training a foundation model on the annotated datasets to generate responses conditioned on specified attribute values. Finally, the AC-SFT model can be further improved by bootstrapping more training data through sampling the model to obtain diverse, high-quality responses for additional training.

\paragraph{Modifications}
In contrast to \citet{dong2023steerlm}, our experiments utilize only the  Open Assistant (OASST) dataset for AC-SFT training rather than multiple datasets. We also scale the attribute labels from OASST to a 0-4 range to match the annotations of \helpsteer dataset. To train the Attribute Prediction Model, we combine the OASST and \helpsteer datasets to predict 9 labels in total. We choose to use the Quality, Humor, Toxicity and Creativity labels from the OASST dataset in addition to the 5 labels from \helpsteer data. After initial exploration, we opt to exclude the bootstrapping step of sampling the AC-SFT model and retraining on its generations, as this provided minimal gains. Finally, instead of the language model
based attribute prediction model in \citet{dong2023steerlm}, we employ a regression model, which we find to work better.
We implement this by taking the last hidden state from the Llama 2 foundation model and adding a regression head on top of it for each attribute. With these modifications, we streamline the \steerlm pipeline while retaining its effectiveness in incorporating rich semantic signals for aligning foundation models. 
\paragraph{Hyperparameters} We train both the Attribute Prediction and Attribute-Conditioned SFT models for 800 steps using a global batch size of 128 (close to 2 epochs) and a constant learning rate of 5e-6 with the AdamW optimizer \citep{loshchilov2017decoupled}. Unless otherwise stated (\textit{e.g.} to show steerability), we set all attributes to 4 at inference time, except creativity, humor, and toxicity which are set to 0.

\subsection{Baseline Models}

\paragraph{SFT} We train a model using only Open Assistant prompts and responses, which is identical to \steerlm minus the attribute labels that we use to condition SFT. We train the model for 800 steps with a global batch size of 128 (close to 2 epochs) and a constant learning rate of 5e-6 using the AdamW optimizer, in line with \steerlm training.

\paragraph{RLHF on Open Source Dataset}
Starting from the above SFT model, we conduct RLHF on HH-RLHF \citep{bai2022training}. We train a reward model for one epoch, and select the checkpoint with lowest validation loss. We then optimize the policy network on the same data using the PPO algorithm \citep{schulman2017proximal}. Following \citet{ouyang2022training, touvron2023llama}, we adopt a global batch size of 512 and a mini-batch size of 64 for each PPO iteration. We set the PPO clip threshold to 0.2, $\mathcal{\beta}$ (KL penalty) to 0.005, and sampling temperature to 1 for rollouts. We employ AdamW optimizer and apply a constant learning rate of 9e-7 with a warmup over the first 10 iterations. We train the model for 800 steps using evaluations on held-out validation prompts for checkpoint selection.

\paragraph{DPO on Open Source Datasets}

We implement Direct Preference Optimization, an efficient substitution of RLHF, following the methodology of \citet{rafailov2023direct}. Initializing with the above SFT model, we train two models with DPO, one using the HH-RLHF dataset, and another using the Open Assistant dataset. We train each model for 1 epoch of its respective dataset, with a KL penalty of 0.2 and a global batch size of 512. We use the AdamW optimizer with a constant learning rate of 9e-6, weight decay 0.1, betas of (0.9, 0.98) with 10 warmup steps.

\paragraph{Llama 2 70B Chat} is a popular RLHF model \citep{touvron2023llama}, which uses the same foundation model as \steerlm but trained with closed-source data. It is trained using 27,540 private SFT samples and then trained with RLHF on 1.4 million private pairs of comparison samples. We use MT Bench score from Chatbot Arena Leaderboard\footnote{\url{https://huggingface.co/spaces/lmsys/chatbot-arena-leaderboard} } and Truthful MC2 score from Open LLM Leaderboard\footnote{\url{https://huggingface.co/spaces/HuggingFaceH4/open_llm_leaderboard}}. Other metrics are calculated based on model responses in Huggingface MT Bench space\footnote{\url{https://huggingface.co/spaces/lmsys/mt-bench/blob/main/data/mt_bench/model_answer/Llama-2-70b-chat.jsonl}}.

    \section{Results}

\begin{table}[ht!]
\centering
\begin{adjustbox}{max width=\columnwidth, scale=1
}
\begin{tabular}{lccccc}
\toprule
\textit{Model} & \textit{MTBench $\uparrow$} &  \textit{TruthfulQA$\uparrow$} & \textit{PPL $\downarrow$} & \textit{FGKL$\uparrow$} & {Chars.}\\

\midrule
\steerlm & \textbf{7.54}& \textbf{0.5613} & \textbf{2.876}& \textbf{8.658}& 1192.7\\
SFT & 6.29 & 0.4930 & 8.199& 7.852 & 604.2\\
Llama 2 Chat & 6.86 & 0.5280 & 4.377 & 7.496 & \textbf{1350.6}\\ 
RLHF w. HH-RLHF & 7.21 & 0.5042 & 3.438 & 7.418 & 831.6\\
DPO w. HH-RLHF & 6.94 & 0.5021 & 8.102 & 7.977 & 787.7 \\
DPO w. OASST& 6.98 & 0.5022 & 7.028 & 7.323 & 834.9 \\ 

\bottomrule
\end{tabular}
\end{adjustbox}
\caption{Automatic evaluation of \steerlm against baseline models trained with open source data and Llama 2 Chat. Higher is better for MT Bench, TruthfulQA and FKGL, and lower is better for PPL.}
\label{tab:auto_eval}
\end{table}

Table \ref{tab:auto_eval} demonstrates that when leveraging \helpsteer, \steerlm produces the most helpful, correct, and coherent responses compared to baseline models that can be trained with open-source datasets as well as Llama 2 70B Chat models. On MT Bench (a measure for helpfulness), \steerlm achieves the top score of 7.54. Baseline models trained with RLHF or DPO on open-source datasets (Open Assistant or HH-RLHF) achieve a maximum of 7.21. This is especially significant given that the best performing model (RLHF w. HH-RLHF) requires 5 times as much compute as \steerlm (see Appendix \ref{sec:compute}), while alternatives requiring similar compute to \steerlm trail further behind (MT Bench $\leq$ 6.98).

\begin{table}[ht!]
\centering
\begin{adjustbox}{max width=\columnwidth, scale=1
}
\begin{tabular}{lccccc}
\toprule
\textit{Model} & \multicolumn{3}{c}{\textit{Win Rate (\%) vs.}} &  \textit{Elo Score} \\
 & \steerlm & Llama 2 & RLHF w. &  \textit{} \\
  &  & Chat & HH-RLHF &  \textit{} \\
\midrule
\steerlm & - & 57.5 & 62.9 & 1050\\
Llama 2 Chat & 42.5 & - & 49.2 & 979\\ 
RLHF w. & 37.1 & 50.8& - & 971\\
HH-RLHF \\

\bottomrule
\end{tabular}
\end{adjustbox}
\caption{Human Evaluation. Higher is better for Win Rate and Elo Score.}
\label{tab:human_eval}
\end{table}

We also conduct human evaluation to complement the automatic evaluation in understanding the relative helpfulness of model responses. Given resource constraints, we were only able to conduct human evaluation on three models - \steerlm, Llama 2 Chat and RLHF w. HH-RLHF. As seen in Table \ref{tab:human_eval}, \steerlm attained the highest Elo rating of 1050 based on pairwise model comparisons, with $57.5\%$ win rate against Llama 2 Chat and $62.9\%$ against our RLHF w. HH-RLHF baseline. 

\begin{figure}[t]
    \centering
    \includegraphics[width=7.6cm]{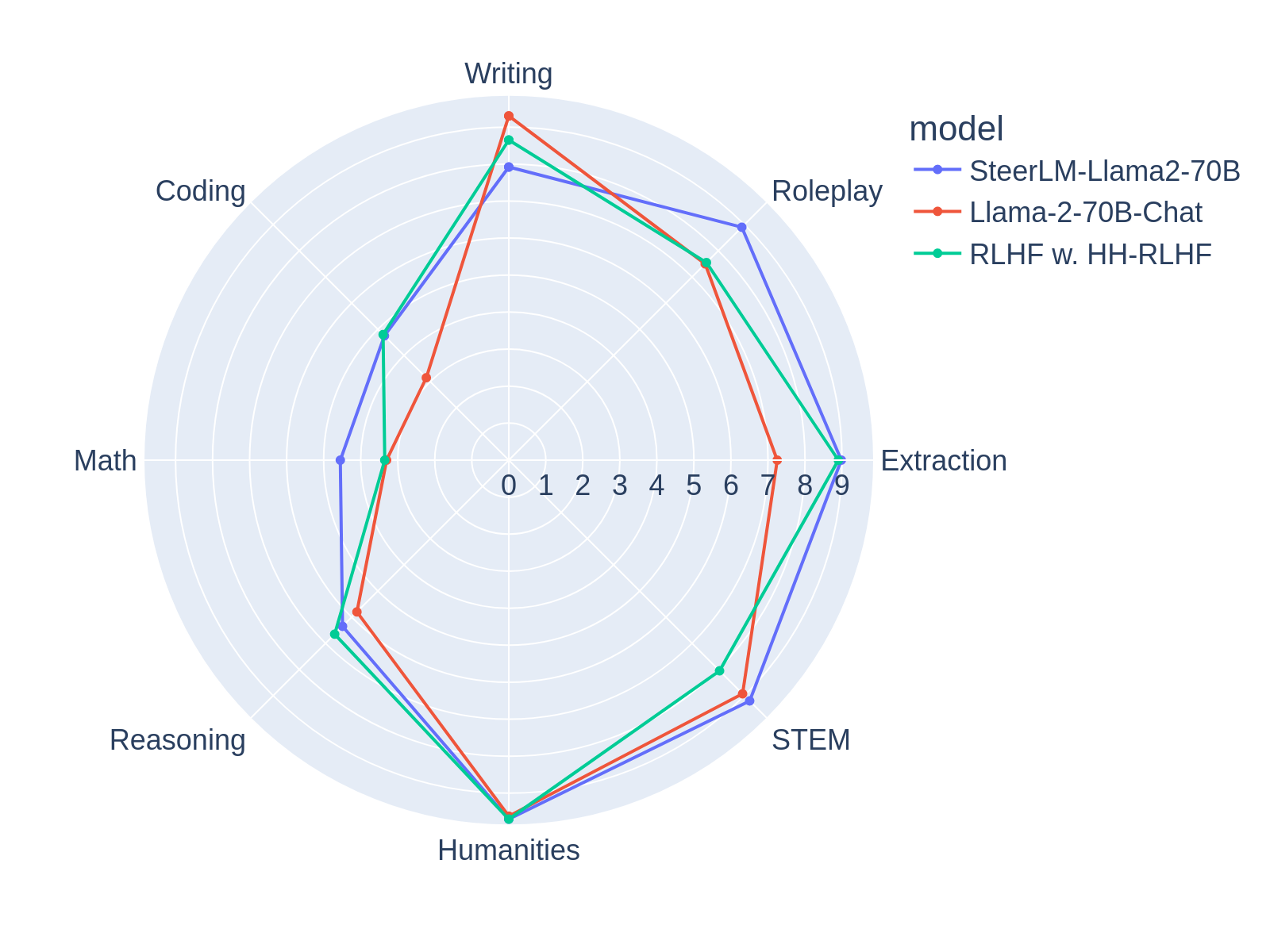}
    \caption{\steerlm performs better than Llama2 Chat and RLHF w. HH-RLHF models in most MT Bench categories.}
    \label{fig:mt_benchmark_break_down}
\end{figure}

When we break down MT Bench performance by category (Figure \ref{fig:mt_benchmark_break_down}), \steerlm model achieves a large gain over Llama 2 70B Chat in categories like \texttt{"Extraction", "Coding", "Math", "Reasoning"} and \texttt{"Roleplay"} for which the model needs to follow instructions precisely and produce correct answers. The large gain can be partially explained by \steerlm's high performance on TruthfulQA (0.5613), good response coherence (2.876 PPL.) and greater complexity of response (FKGL of 8.658). At the same time, \steerlm produces responses with sufficient details (mean length of 1192.7 characters) contrasted with the succinctness of SFT/RLHF/DPO models trained on open-source data and the highly verbose Llama2 Chat. This demonstrates the success of using attributes like correctness, coherence, complexity and verbosity to condition generation.

\section{Ablation Studies}

To better understand the contributions of each \helpsteer attribute, we conduct ablation studies in which we exclude one or more attribute(s) when training the Attribute-Condition SFT model. As seen in Table \ref{tab:ablation_study}, all five attributes within \helpsteer contribute towards improving MT Bench, as MT Bench is lowered when any of them is removed. Furthermore, removing an attribute also lowers the performance on the metrics that measure that attribute, suggesting that \steerlm can effectively use each attribute to condition generation. 

\begin{table}[ht!]
\centering
\begin{adjustbox}{max width=\columnwidth, scale=1
}
\begin{tabular}{lccccc}
\toprule
\textit{Model} & \textit{MTBench$\uparrow$} &  \textit{TruthfulQA$\uparrow$} & \textit{PPL$\downarrow$} & \textit{FGKL$\uparrow$} & {Chars.}\\
\midrule

\steerlm  & \textbf{7.54}	& 0.5613 & \textbf{2.876} & 8.658 & \textbf{1192.7} \\
- helpfulness & 7.17 & \textbf{0.5754} & 3.066 & 8.571 & 1085.6\\
- correctness & 6.92 & 0.5474 & 3.014 & \textbf{8.991} & 1175.4\\
- coherence & 7.13 & 0.5381 & 2.973 & 8.265 & 1170\\
- complexity & 7.12 & 0.5374 & 2.872& 8.019 & 1143.1\\
- verbosity & 7.07 & 0.5217 & 3.718 & 8.333 & 1021.6\\
- \helpsteer & 6.9 & 0.5393	& 6.138 & 7.945 & 825.4\\
- OASST & 7.36 & 0.5557 & 3.244 & 8.483 & 1022.4\\
- \helpsteer -  \\
OASST (=SFT) & 6.29 & 0.493 & 8.199& 7.852 & 604.2 \\

\bottomrule
\end{tabular}
\end{adjustbox}
\caption{Ablation studies with automatic evaluation metrics. Each row represents the performance of \steerlm when the associated attribute(s) are excluded when training the Attribute-Condition SFT model.}
\label{tab:ablation_study}
\end{table}

Among the various attributes, ablations for helpfulness and correctness stand out. The increase in TruthfulQA MC2 when the helpfulness attribute is removed (0.5613 $\rightarrow$ 0.5754) shows that correctness can be further improved when not \textit{explicitly} optimizing for helpfulness. A corollary of this ablation is that optimizing language models for only helpfulness, as done in mainstream RLHF \citep{bai2022training, ouyang2022training, touvron2023llama}, might result in models being sub-optimal in terms of its correctness, as also observed by work on reward model overoptimization \citep{lambert2023alignment}.

 The ablation of correctness shows the importance of the model's factuality to its overall helpfulness. When a model is not explicitly trained to be truthful, it can substantially hurt its overall helpfulness, as shown in the large drop in MT Bench (7.54 $\rightarrow$ 6.92). On the other hand, such a model can generate a more complex-sounding responses, resembling materials for a higher reading age (FKGL). The possibility of language models being optimized to generate complex-sounding responses that are poorly grounded in real world facts re-iteriates the importance of explicitly training models to be factual in addition to being helpful.

 Finally, comparing the ablation of \helpsteer attributes to Open Assistant attributes suggests that \helpsteer attributes contribute more towards the model's helpfulness since its drop in MT Bench is more substantial (7.54 $\rightarrow$ 6.9 vs. 7.54 $\rightarrow$ 7.36). Using both sets of attributes together complements each other, resulting in the most helpful model (MT Bench 7.54) while training with either set of attributes alone can still far outperform a vanilla SFT model (MT Bench 6.29).

\section{Steerability Demo}

A further advantage \steerlm offers is the ability to customize responses at inference time, providing end-users the ability to adapt to different scenarios. 
We present an analysis of steering the attributes of complexity and verbosity, which can be important for different user needs. In educational settings, presenting information with lower complexity aids comprehension for beginning learners. Conversely, for experts, higher complexity can enhance their advanced understanding. When we increase the complexity attribute, the reading grade-level (FKGL) of the response gradually increases from 8.489 to 9.104. 
For instance, when prompted with a question, \texttt{What is gravity?}

A low complexity response looks like

\begin{tcolorbox}
Gravity is the force that pulls everything towards each other. It is what keeps the planets in orbit around the sun, and what causes the tides in the ocean to rise and fall ...
\end{tcolorbox}

while a high complexity one looks like,

\begin{tcolorbox}
Gravity is a force that exists between any two objects with mass or energy, causing them to attract each other. Gravity is a result of the curvature of spacetime caused by the presence of mass or energy ...
\end{tcolorbox}

In business settings, different stakeholders (\textit{e.g.} executives, managers, and engineers) might need answers to similar questions at varying levels of detail. As we dial up verbosity, mean response length increases from 422.3 to 1125.2 characters.

\section{Conclusion}

We create \helpsteer, a multiple attribute helpfulness preference dataset for \steerlm with 37k samples, covering attributes of correctness, coherence, complexity and verbosity in addition to overall helpfulness. 
We openly release this dataset with CC-BY-4.0 license.
Training a Llama 2 70B model on \helpsteer using the \steerlm technique produces a model that scores 7.54 on MT Bench, highest among models trained with open source datasets. Beyond helpfulness, this model is also more truthful and coherent compared to Llama 2 70B Chat and other baselines while being user-steerable in verbosity and complexity.

\section*{Limitations}

\helpsteer contains only annotations for prompts and responses in English. While we did not evaluate \steerlm on multilingual benchmarks, it is unlikely to improve the performance of models on non-English prompts. Nonetheless, our dataset collection methodology can be applied to collect  annotations for a similar dataset in other languages. 

\helpsteer annotations are also likely to reflect what is construed as helpful in the United States since all annotators are based in the US. With the understanding that helpfulness in responses might be culture-specific, we are not certain that this dataset will reflect the opinions of those based in other countries. For these situations, we believe our dataset collection methodology can be used to collect further annotations to be capture helpfulness in other cultures.

\section*{Ethics Statement}

Annotators for the \helpsteer dataset were contracted through Scale AI. Scale AI engages the Anker Methodology, GISC Impact Sourcing Standard, and UN Sustainable Development Goals to provide a fair and competitive pay. The specific pay is calculated based on many factors, including the specific project, the specialized skillset and expertise required, regional costs of living and then transparently listed on Scale AI platform. Scale AI also provides multiple channels for questions and support, including 24/7 support teams, community discussion channels with specially trained moderators, and a “speak up” hotline where contractors can report concerns anonymously. Worker concerns can be submitted to and are reviewed by the Remotasks support team, and pay disputes are reviewed by support specialists trained in this area.

\section*{Acknowledgments} We would like to thank many people at NVIDIA and Scale AI who supported this project. Specifically, we would like to thank Shengyang Sun for reviewing an earlier draft of the paper. 

\bibliography{anthology,custom}

\appendix

\section{Appendix}\label{sec:appendix}

\subsection{\helpsteer Annotation Guidelines}\label{sec:appendix_annotation_guidelines}

\paragraph{Context} NVIDIA is working on creating a Large Language Model (LLM) that can follow instructions and give the appropriate answers. As part of this effort, fine-grained evaluation of the responses given by models and by humans, across different axes and attributes, is fundamental to understanding how improvements and changes are affecting performance. Thus, we ask participants to evaluate prompt-response pairs on several criteria to help the LLM team assess its performance.

\paragraph{Instructions}
You will be given prompts/instructions and a variable number of outputs. Your task is to rate those outputs based on these 5 axes, each on a 5 point likert scale. 

\begin{enumerate}
    \item Helpfulness: How useful and helpful the overall response is.
    \item Correctness: The response is based on facts, no hallucinations, no mistakes. The response covers everything required in the instruction. 
    \item Coherence: The response is self-consistent in terms of content, style of writing, and does not contradict itself. The response can be logically followed and understood by a human. The response does not contain redundant or repeated information.
    \item Complexity: Rate the response along a simple to complex spectrum. A simple response uses simple, easy to understand vocabulary and sentence structure that children can understand. Conversely, a complex response uses sophisticated language with enhanced vocabulary that adults with advanced education or experts on the topic would use. 
    \item Verbosity: A low verbosity response is direct to the point without extra wordings. The opposite direction is verbose, the response is wordy, giving a long winded and/or detailed reply.
\end{enumerate}

Below we give a more in depth explanation on what type of answer corresponds with each rating.

\paragraph{Helpfulness}

\begin{enumerate}
    \setcounter{enumi}{-1}
    \item The response is not useful or helpful at all. The response completely missed the essence of what the user wanted. 
    \item The response is borderline unhelpful and mostly does not capture what the user was looking for, but is still usable and helpful in a small way.
    \item The response is partially helpful but misses the overall goal of the user's query/input in some way. The response did not fully satisfy what the user was looking for.
    \item The response is mostly helpful and mainly aligned with what the user was looking for, but there is still some room for improvement.
    \item The response is extremely helpful and completely aligned with the spirit of what the prompt was asking for.
\end{enumerate}

\paragraph{Correctness}

\begin{enumerate}
    \setcounter{enumi}{-1}
    \item The response is completely incorrect. All information provided is wrong, false or hallucinated. If the prompt asks the assistant to do a task, the task is not at all attempted, or the wrong task was attempted in the response. The response is completely irrelevant to the prompt.
    \item The response has some correct elements but is mostly wrong or incomplete. The response may contain multiple instances of hallucinations, false information, misleading information, or irrelevant information. If the prompt asks the assistant to do a task, the task was attempted with a small amount of success.  
    \item The response contains a mix of correct and incorrect information. The response may miss some details, contain misleading information, or minor hallucinations, but is more or less aligned with what the prompt asks for. If the prompt asks the assistant to perform a task, the task is attempted with moderate success but still has clear room for improvement. 
    \item The response is mostly accurate and correct with a small amount of missing information. It contains no misleading information or hallucinations. If the prompt asks the assistant to perform a task, the task is mostly successfully attempted.
    \item The response is completely correct and accurate to what is requested by the prompt with no necessary details missing and without false, misleading, or hallucinated information. If the prompt asks the assistant to do a task, the task is completely done and addressed in the response.
\end{enumerate}

\paragraph{Coherence}

With this attribute, we measure how lucid, cogent, and self-consistent the model’s response is. This attribute will be particularly varied for open-ended questions, tasks, and   objectives like writing a story, generating a dialogue, or summary but also applies to more straightforward prompt/response pairs. 

\begin{enumerate}
    \setcounter{enumi}{-1}
    \item (Completely Incoherent and/or Unclear) - The response is completely incomprehensible and no clear meaning or sensible message can be discerned from it.
    \item (Mostly Incoherent and/or Unclear) - The response is mostly hard to follow, with inconsistencies, contradictions, confusing logic flow, or unclear language used throughout, but there are some coherent/clear parts.
    \item (A Little Unclear and/or Incoherent) - The response is a little unclear. There are some inconsistencies or contradictions, run on sentences, confusing statements, or hard to follow sections of the response. 
    \item (Mostly Coherent and Clear) - The response is mostly clear and coherent, but there may be one or two places where the wording is confusing or the flow of the response is a little hard to follow. Overall, the response can mostly be followed with a little room for improvement. 
    \item (Perfectly Coherent and Clear) - The response is perfectly clear and self-consistent throughout. There are no contradictory assertions or statements, the writing flows logically, and following the train of thought/story is not challenging. 
\end{enumerate}

\paragraph{Complexity}

\begin{enumerate}
    \setcounter{enumi}{-1}
    \item (Basic) - The response uses very easy to understand language that is clear and completely interpretable by children, adults, and anyone with a functional command of the language. 
    \item (Simple) - The response uses relatively straightforward language and wording, but some schooling through elementary or a middle school in the language might be required to understand the response.
    \item (Intermediate) - People who have completed up through a high school education will probably be able to understand the vocabulary and sentence structure used, but those at the basic level or children might struggle to understand the response.
    \item (Advanced) - The response uses a fairly sophisticated vocabulary and terminology. Someone majoring in this subject at a college or university could have written it and would understand the response. An average adult who does not work or study in this area could not have written the response.
    \item (Expert) - An expert in the field or area could have written the response. It uses specific and technically relevant vocabulary. It contains elevated language that someone at the simple or basic level may not understand at all. The professional language of a lawyer, scientist, engineer, or doctor falls into this category. 
\end{enumerate}

\paragraph{Verbosity} The goal here is to place the response on a spectrum from the most short, crisp  answers, to the most lengthy, detailed, and/or wordy answers under the context of what a user is expecting as a response to the prompt. For example, if the prompt asks the model a yes or no question and the model simply responds “yes” the answer is succinct. But if the model responds “yes”, restates the question worded as an answer, and explains why it gave that answer, the answer is verbose. Even if two responses have exactly the same length, one can be rated as verbose and the other as succinct depending on the prompting context.

\begin{enumerate}
    \setcounter{enumi}{-1}
    \item (Succinct) - The response is short, to the point, and the most concise it can be. No additional information is provided outside of what is requested by the prompt.
    \item (Pretty Short) - The response is on the shorter side but could still have words, details, and/or text removed before it’s at a bare minimum of what the response is trying to convey.
    \item (Average Length) - The response isn’t especially long or short given what the prompt is asking of the model. The length is adequate for conveying a full response but isn’t particularly wordy nor particularly concise. 
    \item (Moderately Long) - The response is on the longer side but could still have more added to it before it is considered fully detailed or rambling.
    \item (Verbose) - The response is particularly lengthy, wordy, and/or extensive with extra details given what the prompt requested from the assistant model. The response can be verbose regardless of if the length is due to repetition and incoherency or if it is due to rich and insightful detail.
\end{enumerate}

\paragraph{Example for annotators} In addition, we give annotators 5 example prompts, each with 2 to 3 responses annotated with \helpsteer attribute labels. We provide one such example prompt below with the sample responses and labels.

\paragraph{Prompt} Blue light is a wavelength that’s naturally present in light, including light from the sun. Compared to other light types, blue light has a short wavelength. This is important because doctors associate short-wavelength light with greater risk for eye damage.
While many electronics (including lightbulbs) emit blue light, computer screens and televisions typically release more blue light than other electronics. This is because computers and televisions commonly use liquid crystal displays or LCDs. These screens may look super-clear and vibrant, but they also release more blue light than their non-LCD counterparts.

Summarize the text into 2 sentences.
\paragraph{Response 1} Blue light is a wavelength that’s naturally present in light, including light from the sun.

\paragraph{Labels 1}
\begin{enumerate}
    \item Helpfulness: 0
    \item Correctness: 1
    \item Coherence: 4
    \item Complexity: 1
    \item Verbosity: 1
\end{enumerate}

\paragraph{Response 2} Blue light is a wavelength that is naturally present in light, and it is associated with greater risk for eye damage. Computer screens and televisions typically release more blue light than other electronics because they use liquid crystal displays.

\paragraph{Labels 2}
\begin{enumerate}
    \item Helpfulness: 4
    \item Correctness: 4
    \item Coherence: 4
    \item Complexity: 3
    \item Verbosity: 2
\end{enumerate}

\paragraph{Response 3} Blue light is a type of light with a short wavelength that is emitted from both natural and artificial sources. Blue light exposure is associated with eye damage, especially from prolonged exposure to artificial sources such as computer screens and televisions.

\paragraph{Labels 3}
\begin{enumerate}
    \item Helpfulness: 4
    \item Correctness: 4
    \item Coherence: 4
    \item Complexity: 2
    \item Verbosity: 2
\end{enumerate}

\subsection{Compute requirements for training models}\label{sec:compute}
\begin{table}[ht!]
\centering
\begin{adjustbox}{max width=\columnwidth, scale=1
}
\begin{tabular}{lccccc}
\toprule
\textit{Model} & \textit{MTBench $\uparrow$} &  Compute (GPU-hours) $\downarrow$  \\
\midrule
\steerlm & 7.54 & 1536  \\
SFT & 6.29 & 1024 \\
RLHF w. HH-RLHF & 7.21 & 7168\\
DPO w. HH-RLHF  & 6.94 & 2048 \\
DPO w. OASST & 6.98 & 1024 \\ 

\bottomrule
\end{tabular}
\end{adjustbox}
\caption{Compute against MT Bench, which measures model helpfulness. Higher is better for MT Bench and lower is better for Compute. Experiments are run on 32 nodes of 8 A100-80GB SXM GPUs each. \citet{touvron2023llama} does not release compute required for performing SFT and RLHF on the Llama 2 70B Chat model.}
\label{tab:compute}
\end{table}

\end{document}